\documentclass[letterpaper, 10 pt, journal, twoside]{IEEEtran}
\usepackage{amsmath,amsfonts}
\usepackage{algorithmic}
\usepackage{algorithm}
\usepackage{array}
\usepackage[caption=false,font=normalsize,labelfont=sf,textfont=sf]{subfig}
\usepackage{textcomp}
\usepackage{stfloats}
\usepackage{url}
\usepackage{verbatim}
\usepackage{graphicx}
\usepackage{cite}
\begin{document}

\title{Diverse Multiple Trajectory Prediction Using a Two-stage Prediction Network Trained with Lane Loss}

\author{Sanmin Kim, Hyeongseok Jeon, Jun Won Choi and Dongsuk Kum
\thanks{Manuscript received: August 2, 2022; Revised: November 9, 2022; Accepted: December 9, 2022.}
\thanks{This paper was recommended for publication by Editor M. Vincze upon evaluation of the Associate Editor and Reviewers’ comments.}
\thanks{This work was supported by Institute of Information \& communications Technology Planning \& Evaluation (IITP) grant funded by the Korea government (MSIT) (No.2021-0-00951, Development of Cloud based Autonomous Driving AI learning Software).}
\thanks{Sanmin Kim, and Dongsuk Kum are with the Graduate School of Mobility, Korea Advanced Institute of Science and Technology (KAIST), 34141 Daejeon, Republic of Korea, (e-mail: sanmin.kim@kaist.ac.kr; dskum@kaist.ac.kr).}
\thanks{Hyeongseok Jeon is with the Advanced Technology R\&D Center, MORAI Inc., 06160 Seoul, Republic of Korea.
(e-mail: hsjeon@morai.ai).}
\thanks{Jun Won Choi is with the Department of Electrical Engineering, Hanyang University, 04763 Seoul, Republic of Korea, (e-mail: junwchoi@hanyang.ac.kr).}
\thanks{Digital Object Identifier (DOI): see top of this page.}
}

\markboth{IEEE ROBOTICS AND AUTOMATION LETTERS. PREPRINT VERSION. ACCEPTED DECEMBER, 2022}%
{Kim \MakeLowercase{\textit{et al.}}: Diverse Multiple Trajectory Prediction using a Two-stage Prediction Network trained with Lane Loss}

\IEEEpubid{0000--0000/00\$00.00~\copyright~2021 IEEE}

\maketitle

\begin{abstract}
Prior studies in the field of motion predictions for autonomous driving tend to focus on finding a trajectory that is close to the ground truth trajectory, which is highly biased toward straight maneuvers. 
Such problem formulations and imbalanced distribution of datasets, however, frequently lead to a loss of diversity and biased trajectory predictions. Therefore, they are unsuitable for real-world autonomous driving, where diverse and road-dependent multimodal trajectory predictions are critical for safety. 
To this end, this study proposes a novel trajectory prediction model that ensures map-adaptive diversity and accommodates geometric constraints.
A two-stage trajectory prediction architecture with a novel trajectory candidate proposal module, \textit{Trajectory Prediction Attention (TPA)}, which is trained with \textit{Lane Loss}, encourages multiple trajectories to be diversely distributed in a map-aware manner. 
Furthermore, the diversity of multiple trajectory predictions cannot be properly evaluated by existing metrics, and thus a novel quantitative evaluation metric, termed the minimum lane final displacement error (minLaneFDE), is also proposed to evaluate the diversity as well as the accuracy of multiple trajectory predictions. 
Experiments conducted on the Argoverse dataset show that the proposed method simultaneously improves the diversity and accuracy of the predicted trajectories.
\end{abstract}


\section{Introduction}

\IEEEPARstart{F}{orecasting}F the future trajectories of dynamic agents in a multi-agent environment is a fundamental task for mobile robotics such as autonomous driving. To take action preemptively for an uncertain future, future motion predictions of nearby agents are necessary. 
However, future motions of nearby agents are highly uncertain, meaning that the deterministic strategy cannot aptly predict their future (e.g., selecting a single trajectory with the most likelihood). 
For this reason, predictions that can cover a wide range of uncertain futures, even those less likely to occur, are required in order for the decision/planning module to guarantee safety at all times so that an autonomous agent can beware of it. 

\begin{figure}[t]
\centering
\includegraphics[width=0.47\textwidth]{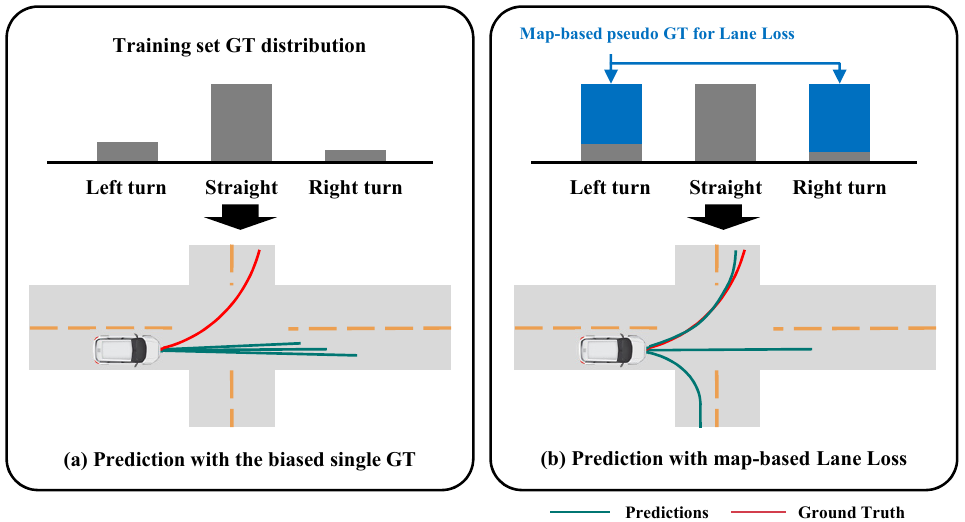}
\vspace{-5pt}
\caption{Importance of diverse predictions. Prediction results with a single ground truth biased toward going straight tend to have a similar distribution to a biased GT distribution. However, our loss term, which leverages map-based pseudo GT, can cover diverse prediction results.} 
\label{figure:1}
\vspace{-15pt}
\end{figure}

Although numerous  trajectory prediction models have been introduced \cite{cslstm, lee2017desire, LaneGCN}, developing a reliable prediction model capable of generating multiple trajectories with diverse maneuvers remains challenging due to several issues. 
First, there is only a single ground truth modality per scenario in the datasets. The lack of diverse ground truth data for a given scenario poses a fundamental problem for multiple-trajectory predictions. 
Second, most trajectory datasets have a severely imbalanced distribution in which the lane-keeping trajectory is highly dominant compared to other maneuvers such as lane changes/turns. 
For example, in the Argoverse motion forecasting dataset \cite{chang2019argoverse}, one of the largest trajectory datasets, more than 90 percent of target agents' maneuvers are going straight, as presented in Table \ref{table:data distribution}.
Such an imbalance in the distribution of datasets leads to biased prediction results. 
Third, a driving environment such as a road map plays a key role in motion predictions, but existing methods cannot sufficiently learn a road map because it is only used as one of the inputs and is not explicitly used for supervision.

Previous works \cite{MTP,zhao2020tnt,covernet} have proposed several multi-modal prediction models with decent prediction accuracy in terms of the final and average displacement errors (FDE and ADE).
However, the multi-modality in these models lacks diversity in the prediction results.
In particular, multi-modal prediction models trained with an imbalanced dataset tend to output multiple trajectories distributed longitudinally, increasing the prediction accuracy in instances of imbalanced validation and test sets even if the outputs are not diverse. (As described in Fig.\ref{figure:1}(a), prediction results are overwhelmed by lane-keeping following the data distribution).
Furthermore, conventional evaluation metrics such as ADE/FDE (Average/Final Displacement Error) encourage these biased results because they only depend on the ground truth in which the majority proceeds straight, and ignoring how diverse the outputs are.

    \begin{table}[!t]
    \renewcommand{\arraystretch}{1.4}
    \setlength{\tabcolsep}{12pt}
    \centering
    \caption{Distribution of Target Agents' Maneuver in Argoverse Trajectory Forecasting Dataset } 
    \label{table:data distribution}
    \begin{tabular}{c | c c }
    \hline
    Maneuver & Training  & Validation\\
    \hline
    Going straight & 191024 (92.75\%) & 34958 (90.70\%) \\
    Left turn & 7860 (3.82\%) & 1880 (4.88\%) \\
    Right turn & 4757 (2.31\%) & 1238 (3.21\%) \\
    Left lane change & 1084 (0.53\%) & 284 (0.74\%) \\
    Right lane change & 1217 (0.59\%) & 184 (0.48\%) \\
    \hline
    \end{tabular}
    \vspace{-15pt}
    \end{table}

In this work, we present a method that improves the diversity of multiple-trajectory predictions in a map-adaptive manner. Our novel Lane Loss approach leverages map information to make multiple trajectory outputs cover diverse maneuvers, including those that have never been observed in the ground truth trajectories. 
First, we find multiple feasible reference lanes from the map information and encourage the outputs to be spread over those reference lanes, which can be used as the pseudo GT (Fig.\ref{figure:1}(b)). 
With Lane Loss, the prediction model can output diverse future trajectories, overcoming the imbalance problem and the limitations of unimodal annotation in trajectory datasets. Thereby, downstream motion planners can reckon with uncertain futures during their planning processes. Moreover, because Lane Loss utilizes map information adaptively, it makes the prediction model generalizable for various road environments. It is important to note that the Lane Loss only works in the training phase, and there is no additional burden in terms of inference.

Furthermore, we propose a two-stage trajectory prediction architecture with Trajectory Proposal Attention (TPA), which generates trajectory candidates inspired by a two-stage object detection framework \cite{fastrcnn, fasterrcnn} that leverages region proposals for more accurate detection results. 
TPA generates trajectory proposals in the middle of the prediction model so that the rest of the network can leverage trajectory proposals to output more accurate final prediction results. TPA not only improves the prediction accuracy but also allows Lane Loss to be applied twice for both trajectory candidates and final predictions to enhance the diversity of the prediction results.

Last but not least, we propose a quantitative evaluation metric, minLaneFDE, that can evaluate both the diversity and feasibility at the same time. Because conventional metrics such as the Average Displacement Error (ADE) and the Final Displacement Error (FDE) only measure errors from the ground truth future trajectory, they are impractical for evaluating the diversity of outputs, especially within an imbalanced dataset. To address this problem, minLaneFDE leverages available map information and measures the minimum error between a set of possible lane candidates (feasible maneuvers) and a set of multiple predicted trajectories.  

To summarize, our main contributions are as follows:

\begin{itemize}
\item We devise a loss function that encourages the multiple outputs to be diverse and feasible considering maneuver candidates based on the map information.
\item A two-stage prediction architecture is proposed, which generates trajectory candidates first and leverages them for the final predictions for accuracy and diversity of multiple predictions.
\item We suggest a quantitative evaluation metric, minLaneFDE, that assesses the quality of diversity in multiple outputs in a map-adaptive manner.
\end{itemize}

\section{Related Work}

\label{sec:Section2}

    \begin{figure*}[t]
    \centering
    \includegraphics[scale=0.95]{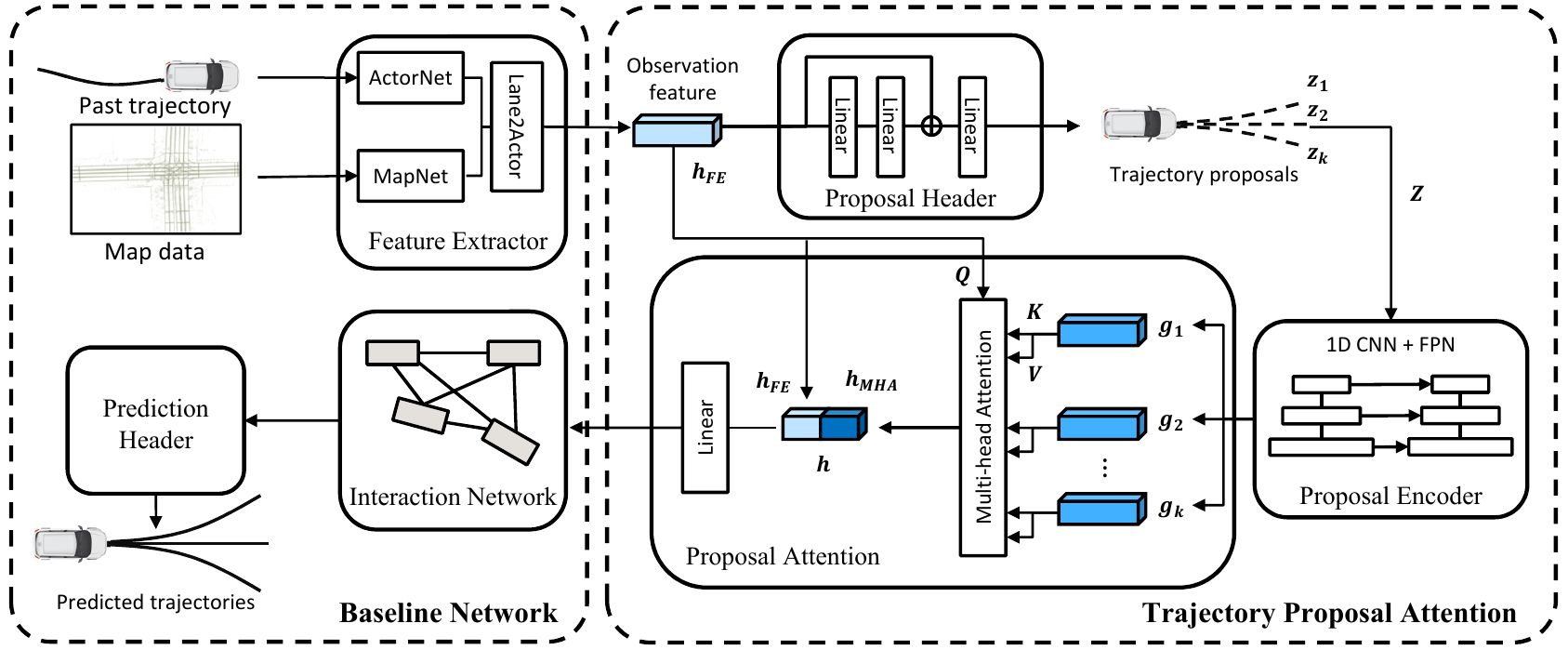}
    \vspace{-8pt}

    \caption{
    The overall architecture of the proposed model consists of the Baseline Network and Trajectory Proposal Attention. Baseline Network describes the general architecture of an interaction-aware prediction model. Trajectory Proposal Attention provides preliminary predictions on the possible future, and both final prediction and proposals are trained with the Lane Loss.
    } 
    \label{figure:2}
    \vspace{-5pt}

    \end{figure*}

\subsection{Multiple Trajectory Prediction}
\noindent 
\textbf{Feature-based multi-modal trajectory regression models}: These models extract features from the past and from the environmental context and then predict multiple trajectories using those features. \cite{MTP2,MATF,MTP} use a simple CNN to encode rasterized map images, while \cite{laneattention,vectornet,LaneGCN,luo2020probabilistic,kim2021lapred} use lane-level map data to handle the complex topology of map data. 
However, they train with only a single modality of the ground truth, underestimating the importance of diversity. 
Despite the fact that several approaches \cite{huang2020diversitygan, yuan2019diverse, rhinehart2018r2p2} have targeted the diversity of outputs, the diversity in these cases does not consider a map \cite{huang2020diversitygan,yuan2019diverse} or only depends on the distribution of the data \cite{rhinehart2018r2p2}, which is highly imbalanced.
 
\vspace{4pt}
\noindent
\textbf{Candidate-based trajectory forecasting approaches}:  \cite{zhao2020tnt,covernet,tpnet,laneattention,lanercnn,mmtransformer} first propose multiple trajectory candidates or goal points, followed by classification and regression to find the most likely candidate. TNT \cite{zhao2020tnt} suggests the goal points candidates of the agent in advance and then undertakes classification and regression for the trajectory to that goal point. On the other hand, CoverNet \cite{covernet} generates a set of trajectories to cover all possible motions before classifying the most likely trajectory among them. Although the previous methods show outstanding performance in terms of prediction accuracy, they are trained with a given ground truth future trajectory representing only a single modality among all possible maneuvers. Therefore, they fail to capture the diversity of predictions. Moreover, these methods are associated with a massive number of candidates, meaning that sufficient solution space must be provided. 

\subsection{Auxiliary Loss for Trajectory Predictions} 
To handle limitations of a distance error-based loss function and to improve the quality of predictions, several auxiliary loss terms \cite{MTP2, messaoud2020multi, niedoba2019improving,boulton2020motion,greer2021trajectory,park2020diverse,casas2020importance, narayanan2021divide} have been introduced.
DAC \cite{narayanan2021divide} proposes a reconstruction loss that encourages two different outputs, one in Cartesian and the other in normal-tangential coordinates, to be equivalent after a coordinate transformation. Although it leverages lane-level map information by representing the trajectory in the curvilinear coordinates of the lane, the loss term targets accuracy and is less sensitive to the degree of diversity.
In one study \cite{greer2021trajectory}, the authors introduce the Yaw loss, which penalizes vehicle trajectories that deviate from the flow of traffic by measuring the angular difference between the vehicle's heading and the nearest lane. 
On the other hand, other studies \cite{casas2020importance} and \cite{niedoba2019improving} introduce auxiliary losses that penalize outputs that lie out of the drivable area.
However, these approaches focus on improving the feasibility and accuracy of trajectories instead of on diversity. 
Although DATF \cite{park2020diverse} introduced a loss term for diverse predictions, it only considers a drivable area, which is less informative.

In contrast to the above methods that attempt to encourage all outputs to be in feasible regions, the proposed Lane Loss method facilitates the multiple outputs so that they are not only feasible but also diverse by minimizing the number of overlapping trajectories.

\section{Problem Formulation}

\label{sec:Section3}
    
We represent the sequences of observed states for agents in the scene at the current time as \(\mathrm{\textbf{S}_{o}} = \{\mathrm{\textbf{S}}^{i}_{o}|_{i=1,\ldots,N}\}\) and \(\mathrm{\textbf{S}}^{i}_{o}=\{\textbf{s}^{i}_{t}|_{t=-t_{o}, -t_{o}+1, \ldots,-1, 0}\}\) where  $i$ is the agent index, $N$ denotes the number of agents, \(\textbf{s}^i_t = \{x^i_t,y^i_t\}\) denotes the 2d coordinate  at time $t$ with the origin at the current position of the agent \(\{x^i_0,y^i_0\}\), and \(t_o\) denotes the observation time horizon.
The set of possible future sequences is denoted as \(\mathrm{\textbf{Y}}=\{\mathrm{\textbf{Y}}^{i,m}|_{{i=1,\ldots,N},{m=1,\ldots,M}}\}\) and \(\mathrm{\textbf{Y}}^{i,m}=\{\mathrm{\textbf{y}}_{t}^{i,m}|_{t=1,2,\ldots,t_f}\} \), where \(\mathrm{\textbf{y}}_t^{i,m} = \{x^{i}_t,y^{i}_t\}\) is a 2d coordinate of the $m$-th modality, and \(M,t_f\) denotes the number of modalities and the prediction time horizon.
Here, $\mathrm{\textbf{Y}}$ includes the annotated ground truth $\mathrm{\textbf{Y}}_{gt}$ because the ground truth is one of the possible future trajectories.
The map information at the current time is represented as \(\mathcal{C}\). We define the whole input as briefly as \(\mathrm{\textbf{X}=\{\textbf{S}_o, \mathcal{C}\}}\).

The objective of the conventional trajectory prediction is to find a model that suitably represents $ p(\mathrm{\textbf{Y}}_{gt}|\mathrm{\textbf{X}}) $ for the given information $\mathrm{\textbf{X}}$. 
In contrast to this approach, our objective is to find a model for  $p(\mathrm{\textbf{Y}}|\mathrm{\textbf{X}})$, where $\mathrm{\textbf{Y}}$ is the set of possible trajectories that denote not only the ground truth $\mathrm{\textbf{Y}}_{gt}$ but also other unseen maneuvers ($\mathrm{\textbf{Y}}_{gt} \in  \mathrm{\textbf{Y}}$). Following this problem definition, the model capable of representing $p(\mathrm{\textbf{Y}}|\mathrm{\textbf{X}})$ is able to cover various motions, including maneuvers that the dataset does not contain as the ground truth. This makes it possible to provide helpful information about uncertain futures to downstream path planners.

\section{Methodology}

\label{sec:Section4}
    
    \subsection{Model Architecture}
    As depicted in Fig. \ref{figure:2}, our model consists mainly of two parts: the Baseline Network and Trajectory Proposal Attention (TPA). In the feature extractor of the baseline network, the historical trajectories of surrounding agents $\textbf{S}_{o}$ are encoded into actor features, and the map information $\mathcal{C}$ is encoded using a Graph Convolutional Network (GCN) \cite{welling2016semi}. The output $\textbf{\textit{h}}_{FE}$, where subscript $\textit{FE}$ denotes the Feature Extractor, is leveraged to generate trajectory proposals and aggregated proposal features in TPA. Then, $\textbf{\textit{h}}$ is fed into the Interaction Network to take into account interactions among agents and maps. Finally, the prediction header outputs the multiple future trajectories of agents and their scores. 
    
    \subsection{Baseline Network}
    The Baseline Network consists of the Feature Extractor, Interaction Network, and Prediction Header. Specifically, we employ LaneGCN \cite{LaneGCN} as our Baseline Network and modify the structure of the model to place the proposed TPA. LaneGCN is a trajectory forecasting model that constructs a lane graph from vectorized map data using a GCN and then fuses the information of agents in the traffic and map features considering the interactions among the agents and maps.
    
    The Feature Extractor consists of ActorNet, MapNet and Lane2Actor, which are sub-modules in LaneGCN \cite{LaneGCN}. ActorNet and MapNet process the trajectories of the agents and map information, respectively, after which Lane2Actor fuses the map and trajectory information. 
    ActorNet is the module for extracting features from the trajectories of agents. This network has the structure of a 1D CNN and a Feature Pyramid Network (FPN) for multi-scale features. 
    MapNet encodes the structured map information into a feature representation. The map data is converted into a graph structure, Lane Graph, in which nodes represent the centerline of the lane segments and the edges represent the connectivity. The multi-scale LaneConv operation to aggregate the topology information of the lane graph is then applied. 

    The Interaction Network considers the interactions among agents and maps using an attention mechanism. Specifically, it takes into account all types of interactions, such as agents-lanes, lanes-lanes, lanes-agents, and agents-agents. 
    
    Lastly, the Prediction Header is the final network for generating multiple future trajectories and confidence scores of each trajectory. Both the regression and scoring network have a residual block and a linear layer.    
    
    \subsection{Trajectory Proposal Attention}
    Trajectory Proposal Attention (TPA) is the first stage of the two-stage trajectory prediction framework. It aims to output multiple trajectory proposals based on historical observations and map data. The proposals are then aggregated into a joint representation $\textit{\textbf{h}}$ through the Proposal Attention. 
    TPA consists of three modules: the Proposal Header, Proposal Encoder, and Proposal Attention.

    \noindent
    \textbf{Proposal Header} takes the feature $\textbf{\textit{h}}_{FE}$ extracted from observable past trajectories and map data as the input and generates trajectory proposals.
    \vspace{-5pt}
    
    \begin{equation} 
    \label{eq:proposal_header}
    \mathrm{\textbf{Z}} = f^{PH}_{\theta}(h_{FE})
    \end{equation}
    
    \noindent    
     where $\mathrm{\textbf{Z}} \in \mathbb{R}^{k \times t_f \times 2}$ is the trajectory proposal, $k$ is the number of proposals, and $\theta$ denotes the parameters for the proposal header. In other words, the proposal header is analogous to a region proposal network (RPN) in two-stage object detection models. The output of the Proposal Header is recursively used in the downstream network to provide physical and geometrical clues. For consistency when generating trajectories among the Proposal Header and Prediction Header, we adopt the same network on both the Proposal Header and Prediction Header. Headers consist of three MLPs with a residual connection after the second layer, a ReLU, and a GroupNorm layer.

    \noindent
    \textbf{Proposal Encoder} is an embedding layer that encodes trajectory proposals into feature vectors.
    \vspace{-5pt}
    
    \begin{equation} 
    \label{eq:proposal_header}
    \textbf{\textit{g}} = \{\textbf{\textit{g}}_{i} = f^{PE}_{\phi}(\mathrm{z}_{i}) |  \mathrm{z}_{i} \in \mathrm{\textbf{Z}}, i = 1, \ldots, k\}
    \end{equation}
    
    \noindent
    where $\textbf{\textit{g}} \in \mathbb{R}^{k \times 128}$ represents the embedding features of trajectory proposals and $\phi$ denotes parameters for the proposal encoder. Because the purpose of the Proposal Encoder is to make embeddings from trajectory-shaped inputs, it has the same architecture as ActorNet in the Baseline Network, consisting of a 1D CNN and a feature pyramid network. The functions to encode each proposal share their weight.

    \noindent
    \textbf{Proposal Attention} is a module that aggregates the features of multiple proposals representing different maneuvers or behaviors. Because no proposals (candidates) can be ignored, a multi-head attention network \cite{vaswani2017attention} is employed. 
    Proposal Attention takes the observation feature $\textbf{\textit{h}}_{FE}$ as the query, and the proposal features $\textbf{\textit{g}}$ are used as the key and the value.
    \vspace{-5pt}

    \begin{equation} 
    \label{eq:proposal_attention2}
    \textbf{\textit{h}}_{MHA} = MHA(Q=\textbf{\textit{h}}_{FE}, K= \textbf{\textit{g}}, V=\textbf{\textit{g}})
    \vspace{-15pt}
    \end{equation}

    \begin{equation} 
    \label{eq:proposal_attention1}
    \textbf{\textit{h}} = \textbf{\textit{h}}_{FE} \oplus \textbf{\textit{h}}_{MHA}
    \end{equation}
    
    \noindent
    where $MHA$ denotes Multi-head Attention \cite{vaswani2017attention}. 
    The attention mechanism, instead of averaging or summation, helps to take more important features among multiple proposals while retaining the other proposals' features.
    Then, each agent's \(\textbf{\textit{h}}_{FE}\) and \(\textbf{\textit{h}}_{MHA}\) are concatenated into \(\textbf{\textit{h}}\). The combined feature \(\textbf{\textit{h}}\) can be interpreted as the feature that takes into account both observable historical information such as past trajectories, and future information such as proposals. \(\textbf{\textit{h}}\) serves as the input to the Interaction Network to estimate the final prediction.
    
    \subsection{Loss Function}
    The proposed model is trained in an end-to-end manner, and we use the summation of three losses: the scoring loss, the regression loss for the final prediction, and the regression loss for proposals.
    \vspace{-15pt}

    \begin{equation} 
    \label{eq:total_loss}
    \mathcal{L}_{total} = \alpha_{score}\mathcal{L}_{score}+\alpha_{pred}\mathcal{L}_{reg}^{pred}+\alpha_{prop}\mathcal{L}_{reg}^{prop}
    \end{equation}
    
    \noindent
    where $\alpha_{score}, \alpha_{pred}$, and $\alpha_{prop}$ are the weighting factors.
    
    To score multiple trajectories, we adopted a hinge loss to maximize the trajectory score closest to the ground truth while suppressing the scores for other modalities, as described in equation \ref{eq:score_loss}.    
    \vspace{-5pt}
    
    \begin{equation} 
    \label{eq:score_loss}
    \mathcal{L}_{score} = \sum_{\:m{\neq}m^*}^{M}{\max{(0, p_{m}+\epsilon-p_{m^{*}}})}
    \end{equation}
    
    \noindent
    where $M$ is the number of predicted trajectories (the number of modalities) for each agent. $p_{m}$ denotes the score output of the $m$-th prediction, $m^{*}$ denotes a modality with the minimum final displacement error, and $\epsilon$ is the margin.
    
    \(\mathcal{L}_{reg}^{*}\) are regression losses for reducing the prediction error, where the superscripts \(prop\) and \(pred\) denote proposals and predictions, respectively. 
    Each regression loss consists of two terms: Winner-Takes-All loss \(\mathcal{L}_{WTA}\) and Lane Loss \(\mathcal{L}_{Lane}\). 

    \noindent
    \textbf{Winner-Takes-All Loss}: The Winner-Takes-All (WTA) loss is widely employed for multiple trajectory prediction \cite{MTP,MTP2,LaneGCN} to overcome the mode collapse problem inherent in the Mixture-of-Experts (ME) loss.
    The WTA loss selects the winner mode based on the Euclidean distance of the position between the ground truth trajectory and the predicted trajectories. The loss is then calculated only for the selected mode (the winner) based on a distance function instead of the weighted sum of all distance metrics across all modalities.
    \vspace{-5pt}
    
    \begin{equation} 
    \label{eq:wta_loss1}
    \mathcal{L}_{WTA} = \min_{m \in M} dist(\hat{Y}^{m}, Y^{gt})
    \end{equation}
    
    \noindent
    where $\hat{Y}^{m}$ denotes the $m$-th predicted trajectory, and $dist(\cdot)$ represents distance metric functions such as RMSE (Root Mean Square Error) or MAE (Mean Absolute Error).
    The WTA loss causes the predicted trajectories to be updated, specifically the one closest to the ground truth, while the scores are updated for all modalities. It allows each prediction output to specialize in typical maneuvers, successfully overcoming the mode collapse problem. \cite{MTP}
    
    However, in order for WTA loss to work successfully, the ground truth trajectories should be uniformly distributed over various maneuvers (e.g., turning or going straight). Because the WTA loss updates only one modality, it is affected considerably by the distribution of the ground truth trajectories. Therefore, the performance of the model trained with the WTA loss and an imbalanced dataset is limited, especially when generating diverse outputs, despite the fact that doing so is critical for safe motion planning.

    Although several works \cite{makansi2019overcoming,rupprecht2017RWTA,thiede2019analyzing,narayanan2021divide} have been proposed to improve the WTA loss, they focused on the problem of the diluted probability density function of the output compared to the ground truth, assuming that the distribution of the ground truth is sufficiently diverse.
    Considering the fact that the ground truth in trajectory datasets is not diverse given that there is a single ground truth per scenario, which is dominated by a lane-keeping maneuver in most cases, those works fail to overcome the fundamental drawback of the WTA loss. To address this problem, we introduce a new loss function, $Lane Loss$, which utilizes the lane-level map information as the pseudo ground truth for the unrevealed modality in the dataset.
    
    \begin{figure}[t]
    \centering
    \includegraphics[scale=0.93]{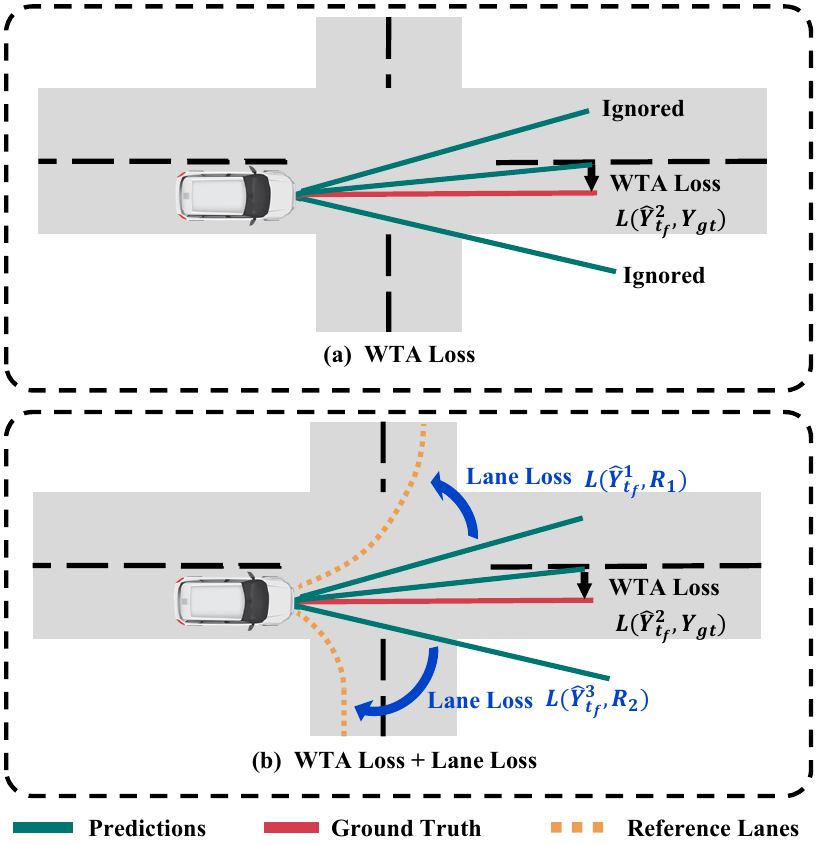}
    \vspace{-5pt}
    \caption{Loss functions for the proposed approach. (a) The Winner-Takes-All (WTA) loss makes the output trajectories close to the ground truth. (b) Lane Loss with the WTA loss makes trajectories close to the ground truth, but the trajectories also follow the reference lanes.}
    \vspace{-10pt}
    \label{figure:3}
    \end{figure}

    \begin{algorithm}[t]
    \caption{Regression loss: Lane loss and WTA loss} 
    \begin{algorithmic}[1]
    
    \STATE\textbf{input}: predicted trajectories \(\hat{Y} \in \mathbb{R}^{M \times t_f \times 2}\), ground truth trajectory \(Y \in \mathbb{R}^{t_f \times 2}\), and reference lanes \(R \in \mathbb{R}^{L \times t_f \times 2}\), where \(M, t_f, L\) denotes the number of prediction outputs (modalities), the prediction time steps, and the number of reference lanes for each  agent, respectively.
    
    \STATE\textbf{output} regression loss $\mathcal{L}_{reg}$
        \FOR {$m$ in $\{1, 2, \ldots, M\}$}
    	    \STATE $ d_{wta}^m \gets$ Euclidean distance between $\hat{Y}^{m}_{tf}$ and $Y_{tf}$
        \ENDFOR
        \STATE $m^* = \arg\min_{m} d_{wta}^{m}$
        \STATE $\mathcal{L}_{WTA} \gets $ Smooth L1 loss between $\hat{Y}^{m^*}$ and $Y$
        \FOR {$l$ in $\{1, 2, \ldots, L\}$}
            \FOR{$m$ in $\{1, 2, \ldots, M\} \backslash m^{*}$}
            \STATE $ d_{lane}^{m} \gets$ Normal dist. of $\hat{Y}^{m}_{tf}$ in Frenet frame of $R_l$
            \ENDFOR
    	    \STATE $m^{**}= \arg\min_{m} d_{lane}^{m}$
    	    \STATE $\mathcal{L}_{lane}^l \gets $ Smooth L1 loss between $\hat{Y}^{m^{**}}$ and $R_l$
    	    \STATE $\mathcal{L}_{lane} \gets \mathcal{L}_{lane} + \mathcal{L}_{lane}^l$
    
        \ENDFOR
        \STATE $\mathcal{L}_{reg} = \mathcal{L}_{WTA} + {1 \over L} \mathcal{L}_{lane}$
        \STATE \textbf{return} $\mathcal{L}_{reg}$
    \end{algorithmic} 
    \label{alg:1}
    \end{algorithm}

    \noindent
    \textbf{Lane Loss}: Despite only a single observed ground truth in each situation, the prediction results should be able to cover other maneuvers that do not align with the maneuver of the ground truth to overcome the limitation of a biased dataset. In other words, it should not ignore the minor portions of situations regardless of the numeric value of the likelihood in order to ensure safe driving in an uncertain future. 
    
    To this end, we introduce Lane Loss (Fig. \ref{figure:3}), with which the multiple trajectories are more likely to be diverse in a map-adaptive manner covering various feasible maneuvers, including cases that are unlikely to occur but are crucial for safety.
    In contrast to other losses, Lane Loss penalizes trajectories not to be identical but to cover possible maneuvers as much as possible.

    Lane Loss uses the distance error of predicted trajectories from the feasible reference lane candidates to encourage diverse predictions. These possible reference lanes are acquired directly from the map data. Similar to the WTA loss, Lane Loss chooses the winner mode for each reference lane candidate based on the Euclidean distance from a lane candidate and the predicted trajectories.     
    \vspace{-5pt}

    \begin{equation} 
    \label{eq:lane_loss}
    \mathcal{L}_{lane} = {\frac{1}{L}}\sum_{l}^{L}\sum_{\:m{\neq}m^*}^{M} u_{m}^{lane} \: {dist}(\hat{Y}_{t_f}^{m}, R_l)
    \end{equation}
    \vspace{-20pt}
    
    \begin{equation}
    \label{eq:lane_loss}
    u_{i}^{lane} = \delta (i = \arg\min_{m} |n_{R_l}(\hat{\mathrm{\textbf{y}}}_{t_f}^m)|)
    \end{equation}
    \vspace{-5pt}
    
    \begin{table*}[t]
    \setlength{\tabcolsep}{3pt}
    \renewcommand{\arraystretch}{1.2}

    \caption{Ablation Study of Modules on Argoverse Motion Forecasting Validation Set} 
    \label{table:ablation}
    \centering
    \resizebox{\textwidth}{!}
    {\begin{tabular}{l || c c c c | c | c c c c | c}
    \hline
     & \multicolumn{5}{c|}{Total} & \multicolumn{5}{c}{Turn subset} \\
    \hline
    Component & minADE$_{6}$ & minFDE$_{6}$ & minADE$_{1}$ & minFDE$_{1}$ & minLaneFDE$_{6}$ & minADE$_{6}$ & minFDE$_{6}$ & minADE$_{1}$ & minFDE$_{1}$ & minLaneFDE$_{6}$ \\
    \hline
    Baseline &\textbf{0.71} & 1.08 & 1.35 & 2.97 & 2.65 & 1.08 & 2.16 & 2.16 & 5.30 & 2.12\\
     + TPA & \textbf{0.71} & \textbf{1.07} & \textbf{1.33} & 2.93 & 2.57 & 1.07 & 2.08 & 2.13 & 5.22 & 1.88\\
     + Lane Loss & 0.73 & 1.13 & 1.36 & 2.98 & 1.56 & 1.06 & 2.04 & 2.11 & 5.05 &  1.35\\
     + Lane Loss + TPA & 0.72 & 1.12 & \textbf{1.33} & \textbf{2.91} & \textbf{1.53} & \textbf{1.05} & \textbf{2.01} & \textbf{2.08} & \textbf{5.02} & \textbf{1.24}\\
    \hline
    
    \end{tabular}}
    \vspace{-10pt}
    \end{table*}

    \noindent
    where $L$ is the number of the reference lanes; $R_l \in \mathbb{R}^{t_f \times 2}$ denotes the $l$-th reference lane with a set of 2d points; \(\delta\) denotes the Kronecker delta, which returns 1 if the condition is true or 0 otherwise; and \(|n_{R_{l}}(\hat{\mathrm{\textbf{y}}}_{t_f}^m)|\) refers to the normal distance of the \(m\)-th predicted trajectory at the final prediction time \(t_{f}\) from the \(l\)-th reference lane \(R_{l}\) in the Frenet-Serret frame. 
    
    The reference lanes for Lane Loss are extracted from the lane graph using the Argoverse API \cite{WIMP}. Specifically, first we find near lane segments in the lane graph, which are located around the vehicle's current position. Then, we expand the selected lane segments using the lane graph, remove overlapping lanes, and finally choose the top-k lanes based on the scores. The details are reported in \cite{WIMP}.
        
    Consequently, the regression loss is the mean of the summation of the WTA loss and Lane Loss for all agents in the scene. 
    \vspace{-10pt}

    \begin{equation} 
    \label{eq:reg_loss}
    \mathcal{L}_{reg}^{*} = {\frac{1}{N}} \sum_{i}^{N} (\mathcal{L}_{WTA}^{i} + \mathcal{L}_{Lane}^{i})
    \end{equation}

    \noindent
    where $\mathcal{L}_{WTA}^{i}$ and $\mathcal{L}_{lane}^{i}$ denotes losses for $i$-th agent. 
    Algorithm \ref{alg:1} presents the pseudocode for the  regression loss.

\section{Experiments}

\label{sec:Section5}

\subsection{Dataset}
    The Argoverse motion forecasting dataset \cite{chang2019argoverse} is a real-world driving trajectory dataset with over 324k sequences collected in the USA. Each sequence consists of five seconds of a time horizon (two seconds for the past trajectory and three seconds for the future trajectory) with a sampling rate of 10Hz. It also provides the locations of the centroid of the target and surrounding agents and the HD-map data.

    \subsection{Metrics}
    We employ the minimum Average/Final Displacement Error (minADE/minFDE) as the evaluation metrics \cite{chang2019argoverse}.

    \begin{equation}
        minADE_{k} = \min_{m=\{1,2, \ldots, k\}} \cfrac{1}{t_f}\sum_{t}^{t_f} \| \hat{\mathrm{y}}_{t}^{m} - \mathrm{y}_{t} \|_2
    \end{equation}
    \begin{equation}
        minFDE_{k} = \min_{m=\{1,2, \ldots, k\}} \| \hat{\mathrm{y}}_{t_f}^{m} - \mathrm{y}_{t_f} \|_2
    \end{equation}
    
    \noindent
    where $k$ denotes the number of modalities that are the most probable trajectories according to the estimated scores. $minADE$ and $minFDE$ are the average of all target agents in the dataset.

    Although these metrics are widely used in the motion forecasting task, minADE and minFDE only depend on the error from the ground truth; therefore, it is impractical to evaluate whether the predictions are realistic. Several works have thus proposed new metrics in addition to ADE/FDE to measure the feasibility of the outputs.
    The Off-Road Rate \cite{niedoba2019improving} evaluates the feasibility of multiple outputs by calculating the percentage of the outputs that lie out of drivable area, while the Off-Yaw Rate \cite{greer2021trajectory} measures the proportion of trajectories that have different heading angles with the nearest lane. Although both of these metrics leverage the map information for evaluating the quality of the output trajectories, they are not suitable if used to evaluate the diversity of multiple outputs.
    
    In relation to this, we define a new performance metric, $minLaneFDE$, which captures both the quantity and quality of the diversity of multiple outputs based on the map data, specifically the centerlines of the reference lanes. 
    Though earlier work \cite{park2020diverse} proposed a metric for the diversity, only drivable areas are considered, while the proposed minLaneFDE focuses on evaluating whether outputs are covering possible future maneuvers. minLaneFDE denotes the minimum value among the lateral displacement errors between the centerlines of possible lane candidates (we employ the Argoverse map API \cite{chang2019argoverse} to determine the lane candidates) and each multiple predicted trajectory. In other words, minLaneFDE evaluates the error of the closest prediction from each lane candidate.
    
    \begin{equation} 
    \label{eq:laneFDE}
    min \: Lane FDE_{k} = {\frac{1}{L}} \sum_{l}^{L} \min_{m=\{1,\ldots,k\}} |n_{R_{l}}(\hat{\mathrm{\textbf{y}}}_{t_f}^{m})|
    \end{equation}
    
    \noindent
    where $L$ is the number of reference lanes obtained from the map considering the agent's current position, $k$ is the number of estimated trajectories, and $t_f$ denotes the final time index. $|n_{R_{l}}(\hat{\mathrm{\textbf{y}}}_{t_f}^{m})|$ denotes the normal distance of the $m$-th prediction $\hat{\mathrm{\textbf{y}}}_{t_f}^{m}$ in the Frenet-Serret frame of the $l$-th reference lane $R_{l}$.
    
    \begin{table}[!t]
    \centering
    \setlength{\tabcolsep}{5pt}
    \caption{Result on the Argoverse Motion Forecasting Test Set} 
    \label{table:testset}
    \begin{tabular}{l | c || c c c c}
    \hline
    Model & Venue & ADE$_{6}$ & FDE$_{6}$ & ADE$_{1}$ & FDE$_{1}$ \\
    \hline
    Argoverse Baseline & - & 1.71 & 3.29 & 3.45 & 7.88 \\
    TNT \cite{zhao2020tnt}& CoRL 20 & 0.91 & 1.45 & 2.17 & 4.95 \\
    LaneGCN \cite{LaneGCN}& ECCV 20 & \textbf{0.87} & 1.36 & 1.71 & 3.78 \\
    HOME \cite{gilles2021home}& ITSC 21 & 0.92 & 1.36 & 1.72 & 3.73 \\
    LaneRCNN \cite{lanercnn}& IROS 21 & 0.90 & 1.45 & 1.69 & 3.69 \\
    mmTransformer \cite{mmtransformer} & CVPR 21 & \textbf{0.87} & \textbf{1.34} & 1.77 & 4.00\\
    PRIME \cite{PRIME}& CoRL 22 & 1.22 & 1.56 & 1.91 & 3.82\\ 
    AutoBot \cite{autobot} & ICLR 22 & 0.88 & 1.37 & 1.84 & 4.11 \\
    
    \hline
    TPA & - & \textbf{0.87} & 1.35 & 1.68 & 3.69\\
    TPA+Lane Loss & - & 0.89 & 1.40 & \textbf{1.67} & \textbf{3.68}\\
    \hline
    \end{tabular}
    \end{table}

\subsection{Implementation Details}
    The time step of the past trajectory \(t_o\) is 20 at 10Hz (two seconds) and the time step for the future trajectory \(t_f\) is 30 at 10Hz (three seconds). We set the number of proposals and the predictions to 6 ($M=6$). For Lane Loss and minLaneFDE, we use a maximum of three reference lanes ($L=3$). Lastly, as the weighting factors for each loss function, $\alpha_{score} = 1.0, \alpha_{pred} = 1.0,\alpha_{prop} = 0.1$ is used.
    
    For training on the Argoverse Motion Forecasting dataset, we implement the models using Pytorch \cite{paszke2017automatic} and use the Adam optimizer with a batch size of 128. We initialize the training with the learning rate set to 0.001 and decrease it by a factor of 10 after 32 epochs. Although we employ an existing model as the baseline, we did not use a pre-trained model, and we trained the entire model. We train the model on four Nvidia RTX 3090 GPUs for a total of 40 epochs.

    \begin{figure*}[t]
    \centering
    \includegraphics[scale=0.95]{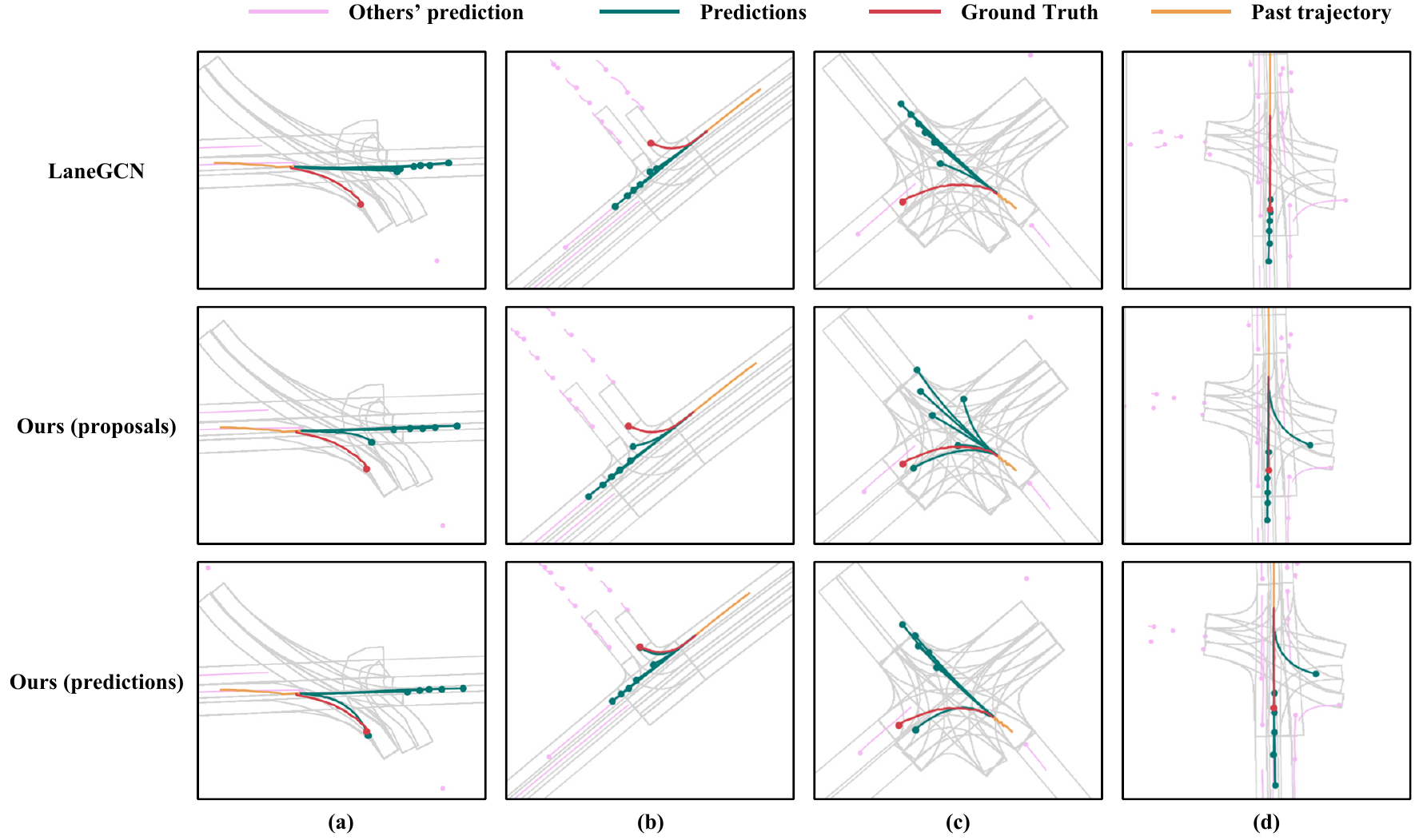}
    \caption{ Visualization of prediction results. \textbf{Top}: Results of LaneGCN. LaneGCN fails to cover diverse trajectories and returns all multiple similar trajectories. \textbf{Middle}: Prediction results of trajectory proposals of our proposed model. Distinct trajectories represent not only going straight even if they are not accurate. \textbf{Bottom}: Final prediction results of our model. Our model can predict accurate and diverse multiple future trajectories.}
    \label{figure:4}
    \end{figure*}
    
    \subsection{Ablation Study}
    To demonstrate the contributions of each component of our model, the results of ablation studies are summarized in Table \ref{table:ablation}. 
    For the total validation set, the model with TPA shows some improvement in all metrics and proves the effectiveness of the two-stage model. Although the model trained with Lane Loss shows outstanding performance in the diversity metric of LaneFDE, it results in inferior predictions in ADE/FDE.
    We claim that the reason for the increase in ADE/FDE comes from the imbalanced distribution in the total validation set. Because going straight is dominant in the validation set, un-diverse outputs which focus on going straight can achieve better performance than diverse outputs which contain other maneuvers.    
    
    To verify this, we conducted a test on a subset of validation that only contains turn maneuvers (the right half of Table \ref{table:ablation}). As a result, the model with Lane Loss shows superior performance even in ADE/FDE metrics, unlike in the total validation set. Moreover, jointly leveraging TPA and Lane Loss brings further improvement in all metrics and achieves the best performance. Conclusively, by training the network with TPA and Lane Loss together, the prediction model can generate diverse outputs in a map-adaptive manner to realize an outstanding advance in both diversity and accuracy.

    \subsection{Comparison with state-of-the-art methods}
     We compare our model with state-of-the-art methods on the test set of the Argoverse Motion Forecasting dataset. These results are presented in Table \ref{table:testset} (``min" is dropped due to the lack of space). It includes the Argoverse official baseline and other candidate-based trajectory prediction approaches similar to the proposed architecture. The results show that the proposed model outperforms the other models. Even if the performance margin is insignificant, considering that our model improves the diversity of prediction output remarkably, this represents an outstanding achievement.

    \subsection{Qualitative Results}
    We visualize the qualitative results of several cases on the Argoverse validation set compared with the baseline model LaneGCN in Figure \ref{figure:4}. 
    The gray lines represent road lanes around the target vehicle, the yellow line denotes the observed past trajectory of the target, the green lines are multiple predictions, which are the output of the model. The red line is the ground truth future trajectory, and the magenta lines are other agents' future trajectories.
    Three rows are the prediction results of LaneGCN, the proposals formulated here (first-stage outputs), and the final prediction of our model.

    As shown in Figure \ref{figure:4}, it is hardly possible to distinguish whether the target agent will go straight or make a turn with the given past trajectory (yellow). In these cases (Fig. \ref{figure:4} (a),(b), and (c)), the baseline model tends to predict the future motion of the target as going straight while ignoring the possibility of other maneuvers. 
    Therefore, it fails to predict the future motion of the agent correctly. 
    However, our model can suggest both straight and turning maneuvers, including the correct prediction. 
    Specifically, as can be verified from the results of the proposals (middle), the proposed model can predict diverse future motions from the stage of trajectory proposals even if they are not accurate compared to the final prediction.
    Furthermore, our model also can cover diverse maneuvers despite the fact that the ground truth is going straight, as in the case shown in Fig. \ref{figure:4}(d).

\section{Conclusion}

\label{sec:Section6}

In this work, we propose a two-stage trajectory prediction framework based on Trajectory Proposal Attention (TPA) which is trained with a novel loss function to encourage multi-modal prediction outputs to be dispersed, leveraging the reference lanes obtained from map data in order to cover diverse maneuvers. The proposed model demonstrated improved performance in both minADE/minFDE and the proposed metric minLaneFDE, which measures the diversity of outputs. In conclusion, our model can provide valuable information on uncertain futures so that decision/planning can proactively take place. Furthermore, because this approach works in a map-adaptive manner, it is applicable to a continuously changing driving environment.

\bibliographystyle{IEEEtran}
\bibliography{root}
\addtolength{\textheight}{-12cm}

\vfill

\end{document}